\pdfoutput=1

\PassOptionsToPackage{numbers, compress}{natbib}

\documentclass[11pt]{article}

\usepackage{ACL2023}

\usepackage{times}
\usepackage{latexsym}
\usepackage{hyperref}
\hypersetup{
    colorlinks=true,
    }
    
\usepackage{graphicx}
\usepackage{subcaption}
\usepackage{multirow}
\usepackage{xcolor}         
\usepackage{float}

\usepackage[T1]{fontenc}

\usepackage[utf8]{inputenc}

\usepackage{microtype}
\usepackage{booktabs}       

\usepackage{inconsolata}

\newcommand{\datalink}{\href{https://www.kaggle.com/datasets/googleai/fleurs-asl}{this link}}

\title{FLEURS-ASL: Including American Sign Language in Massively Multilingual Multitask Evaluation}

\author{Garrett Tanzer \\
  Google \\
  \texttt{gtanzer@google.com}}

\begin{document}
\maketitle
\begin{abstract}
Sign language translation has historically been peripheral to mainstream machine translation research. In order to help converge the fields, we introduce FLEURS-ASL, an extension of the multiway parallel benchmarks FLORES (for text) and FLEURS (for speech) to support their first sign language (as video), American Sign Language, translated by 5 Certified Deaf Interpreters. FLEURS-ASL can be used to evaluate a variety of tasks---primarily sentence- and discourse-level translation---between ASL and 200 other languages as text, or 102 languages as speech. We provide baselines for tasks from ASL to English text using a unified modeling approach that incorporates timestamp tokens and previous text tokens in a 34-second context window, trained on random video clips from YouTube-ASL. This model meets or exceeds the performance of phrase-level baselines while supporting a multitude of new tasks. We also use FLEURS-ASL to show that multimodal frontier models have virtually no understanding of ASL, underscoring the importance of including sign languages in standard evaluation suites.\footnote{We publicly release FLEURS-ASL at {\datalink} under CC BY-SA 4.0.
}
\end{abstract}

\section{Introduction}

Despite calls to include sign languages in mainstream natural language processing~\citep{kayo}, sign language translation remains peripheral to the machine translation field. The sign language translation field uses individual benchmarks for each language---such as How2Sign~\citep{how2sign} for ASL---constructed as train/test splits from one underlying dataset, often with overlapping signers. Meanwhile, mainstream machine translation has developed independently constructed massively multilingual benchmarks like FLORES~\citep{flores2,flores101,flores200} (a set of short documents translated into 200 languages), which has been extended to FLEURS~\citep{fleurs} (speech for 102 languages) and Belebele~\citep{bandarkar2023belebele} (reading comprehension questions for 122 languages). Multiway benchmarks admit better comparisons of translation quality across languages because they reduce irrelevant variation in topic or style. Independently constructed benchmarks give a more representative view of real-world performance because they test out-of-domain generalization rather than held-out generalization; fresh queries from users do not necessarily come from the training distribution.

While YouTube-ASL~\citep{youtubeasl} takes a first step towards this kind of robust evaluation by reporting zero-shot scores on How2Sign (i.e., scores without finetuning on How2Sign), this test set (like others in sign language translation) is limited in many ways beyond just lack of standardization. For example, the content is translated live after only one look-over, by a combination of Hearing interpreters and Deaf signers (not specified to be professional Deaf interpreters), with errors and redos included in-stream but skipped with sentence-level clip boundaries (complicating discourse-level evaluation)~\citep{how2sign,reconsideringsentence}. Of course, zero-shot evaluation on a held-out test split also forces researchers to discard valuable training data.

In this paper, we introduce FLEURS-ASL, an extension of FLORES/FLEURS to support their first sign language (as video), American Sign Language. FLEURS-ASL can be used to evaluate a variety of tasks---primarily sentence- and discourse-level translation, but also in principle caption alignment, retrieval, and multiple-choice receptive comprehension---between ASL and all the languages supported by FLORES and FLEURS. FLEURS-ASL was translated by 5 Certified Deaf Interpreters with an estimated 5-6 hours of preparation per 1 hour of translated content, ensuring a high quality bar.

We provide several sets of baselines. First, we provide human baselines for sentence-level and discourse-level translation, at 13.0 BLEU (64.6 BLEURT) and 13.5 BLEU respectively.

Second, as a model baseline for the suite of FLEURS-ASL tasks, we introduce a unified sign language to text modeling approach that builds upon YouTube-ASL~\citep{youtubeasl}'s T5 baseline. Inspired by Whisper's approach for speech to text tasks~\citep{whisper}, we extend the context window to 34 seconds of signing, include 256 tokens of prior text context, incorporate timestamp tokens as both input and output, and train on random video clips. The use of multi-caption random clips heeds the call of~\citet{reconsideringsentence} to explore methods that can incorporate more context and are more tolerant to caption misalignment. This exploratory method meets or exceeds the performance of sentence-level baselines (3.7 vs. 2.9 BLEU, 37.2 vs. 33.6 BLEURT) while enabling a multitude of new tasks through chunked autoregression, such as discourse-level translation, timed translation, and (albeit poorly) caption alignment---despite lack of optimization due to the large possibility space. 

And third, we use FLEURS-ASL to show that recent multimodal frontier models such as GPT-4o, Claude 3 Opus, and Gemini 1.5 Pro display virtually no understanding of ASL, underscoring the importance of including sign language tasks in standard evaluation suites.

We publicly release FLEURS-ASL at {\datalink}. We hope that it will encourage development and help to evaluate more generally capable sign language models, perhaps drawing inspiration from our new unified modeling approach. We also hope that the practices and findings from our data collection will aid in the construction of high-quality eval sets for more of the world's sign languages. See Section~\ref{sec:limitations} for a discussion of limitations; while there is great value to standardized benchmarks, we expect that more evaluation will be needed specifically for sign languages to capture variety across vision and language aspects and ensure consistent performance.

\section{Related Work}

\subsection{Sign Language Translation Benchmarks}

Sign language translation has historically been evaluated on test splits of the same underlying dataset used for training. Early and well-worn datasets like RWTH-PHOENIX-Weather 2014 T~\citep{slt_orig} have narrow domains---which were initially useful for measuring modeling progress but are not representative of broader content---and consist of live hearing interpretations. Over time datasets have expanded to cover more diverse topics, levels of signing proficiency, recording environment, and so on,~\citep{bobsl,hanke-etal-2020-extending,wmt_slt,gueuwou2023jwsignhighlymultilingualcorpus,joshi2023isltranslatedatasettranslatingindian,wmt_slt_23}, but evaluation still tends to be performed on held-out test splits (often even with overlapping signers). This is true for the most related prior work, How2Sign~\citep{how2sign}, the canonical ASL to English translation benchmark. Beyond the nature of the test split, How2Sign has several issues with translation quality (live interpretation with quality that varies significantly between signers)~\citep{reconsideringsentence}, though it remains valuable due to its unique domain of ``how to'' instructional videos.

The notable exception to this evaluation trend is YouTube-ASL~\citep{youtubeasl} (and the subsequent works that use it~\citep{rust2024privacyaware,gueuwou2024signmusketeersefficientmultistreamapproach}); YouTube-ASL provides no test split but instead evaluates zero-shot and finetuned results on How2Sign.\footnote{YouTube data is also unsuitable as a benchmark because videos may be deleted over time.} Our goal for FLEURS-ASL is to enhance this kind of evaluation across a wide array of tasks, with a focus on quality over quantity.

\subsection{Machine Translation Benchmarks}

In mainstream machine translation, this shift came with benchmarks like FLORES~\citep{flores2,flores101,flores200}, an independently constructed, massively multilingual multiway translation benchmark for over 200 languages as text, which has been extended to FLEURS~\citep{fleurs} for 102 languages as speech, Belebele~\citep{bandarkar2023belebele} with reading comprehension questions for 122 languages, and more through the \href{https://github.com/openlanguagedata/flores}{FLORES+} open source project.

These are by no means the only machine translation benchmarks---there is still space for benchmarks serving other purposes, such as avoiding contamination~\citep{bapna2022buildingmachinetranslationsystems}, hosting fair \href{https://www2.statmt.org/wmt23/}{WMT} competitions in various domains, exercising particular grammatical~\citep{bawden-etal-2018-evaluating} or social aspects~\citep{stanovsky2019evaluatinggenderbiasmachine,costajussa2022evaluatinggenderbiasspeech}, and testing extreme new model capabilities~\citep{tanzer2024benchmarklearningtranslatenew}, among others---but FLORES/FLEURS serve as a valuable foundation for evaluating breadth and robustness of coverage in multilingual models, measured across languages on the same (independent) domain.

\begin{figure*}[t]
    \centering
    \includegraphics[width=\textwidth]{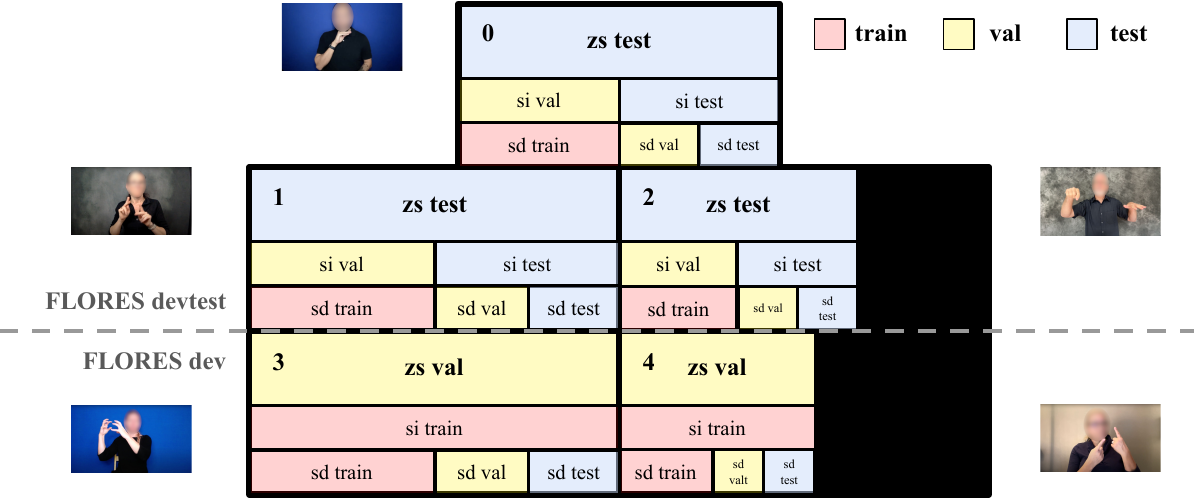} 
    \caption{\textbf{FLEURS-ASL dataset splits.} The sentences are divided among 5 interpreters, and 3 sets of splits: zero-shot (``zs''), signer-independent finetuning (``si''), and signer-dependent finetuning (``sd''). We blur the interpreters' faces in this paper for privacy, but the underlying dataset is unblurred because facial expressions are an essential component of the grammar of sign languages.}
    \label{fig:splits}
\end{figure*}

\subsection{Sign Language Translation Methods}
The bulk of work on methods for sign language translation focuses on alternative featurization, architectures, pretraining, or losses for the same underlying sentence-level translation task framing~\citep{yin2020bettersignlanguagetranslation,moryossef2021evaluatingimmediateapplicabilitypose,de-coster-etal-2021-frozen,zhang2023sltunet,openasl,gueuwou2024signmusketeersefficientmultistreamapproach}. 

To the best of our knowledge, the only prior work that has trained models for context-aware sign language translation is~\citet{sincan2023context}, which incorporates prior text or sign spottings as conditioning for sentence-level translation; the training and test examples still rely on sentence-level alignments being available. BOBSL~\citep{bobsl} and WMT-SLT 23~\citep{wmt_slt_23} include large amounts of weakly supervised data that to our knowledge have not been specially utilized.~\citet{reconsideringsentence} argues based on linguistic survey and an empirical case study that isolated sentence-level clips of the kind in How2Sign are often not even comprehensible to human signers; this motivates our new longer-context modeling approach where we train a mixture of tasks, including conditioning on context and generating timed tracks as outputs, on random crops\footnote{Naturally, works that perform self-supervised pretraining on uncaptioned videos already use random clips.~\citep{hu2021signbertpretraininghandmodelawarerepresentation}} of videos with captions that are generally well-aligned but do not strictly contain the corresponding signing.

\section{The FLEURS-ASL Benchmark}
\label{sec:fleurs-asl}

The FLORES benchmark consists of 3001 sentences across 842 articles drawn from English Wikipedia, split into a dev set of 997 sentences (281 articles), devtest set of 1012 sentences (281 articles), and test set of 992 sentences (280 articles). However, they do not release the test set so---like FLEURS---we build on only the dev and devtest sets. We randomly split these sets into halves (4 chunks in total), each intended to be translated by a different interpreter. This strikes a compromise between signer diversity and the feasibility of recruiting/onboarding qualified interpreters.

\subsection{Data Collection}
\label{sec:data-collection}

We set out specifically to produce translations with a high quality bar, which meant recruiting \href{https://www.casli.org/certified-deaf-interpreter-exam-cdi/}{Certified Deaf Interpreters} (CDIs: Deaf or Hard of Hearing signers with native or near-native fluency, cultural competence, and professional certification for their interpreting qualifications) who would research and plan translations with substantial preparation time, then perform and record them (with a ratio of 5-6 hours of prep time for each 1 hour of recorded content). Therefore our selection criteria were as follows: a) \textit{required}: CDI, b) \textit{required}: expertise in translation from English text,\footnote{Much of the work that CDIs do is actually translating from a Hearing interpreter's ASL into more natural, culturally aware Deaf ASL (with particular focus on facial expressions and body language), so translating the kind of English text found in FLORES is not necessarily within everyone's area of expertise.} c) \textit{required}: has their own professional recording environment, and d) \textit{preferred}: not prolific on YouTube (to maintain zero-shotness).\footnote{Unfortunately, it could not be a hard requirement that the interpreter is completely absent from public web data because the pool of qualified interpreters, especially those who are comfortable interpreting content that will be recorded and made public, is limited.}

We ran an initial pilot through a sign language interpretation vendor, which resulted in signer \#0's chunk. Then we rerandomized the chunks\footnote{We had assumed that the FLORES articles were presented in random order but realized that this is not the case (though the effect seems minor). This rerandomization means that signer \#0's translated content overlaps with signer \#1 and \#2, which may be useful for studying generalization across signers or the variation in translations by different people.} and partnered with the \href{https://dpan.tv/}{Deaf Professional Arts Network (DPAN)}, a Deaf-led organization that specializes in American Sign Language media, to recruit 4 more interpreters (signers \#1-4). Across both of these vendors, candidates were recruited targeted to the above criteria and then screened based on (paid) sample translations they produced for one article. For each candidate that was ultimately selected, several others were considered who either withdrew for logistical reasons or were not the best fit to translate this content.\footnote{In addition to the fact that not all CDIs have expertise in translation from English text, the content of FLORES is relatively difficult to translate because it has many relatively short articles in different domains. Sign language interpreters are often specialized to certain domains or prepare for a domain ahead of an event---but these events are much longer than the short documents being translated here, so the overhead is less burdensome. For text translation benchmarks, different domains can be allocated to different translators because the differences between individual translators are less visible, whereas here we want to be able to study variation across signers without interference from domain correlations.} At a high level, the interpreters were told the project's expectations for the amount of preparation and asked to sign as naturally as possible, flowing naturally between sentences (but preserving the original ordering, which unfortunately limits discourse phenomena to some extent) so that caption alignment wouldn't be artificially easy.\footnote{Despite this instruction, there are sometimes still unnaturally long pauses between sentences due to the performance/memory characteristics of sign language translation, so we do not expect caption alignment to be as challenging as in real data.} See Appendix~\ref{app:translation-guidance} for more details on the guidance interpreters were given. They then proceeded to translate their content. All interpreters gave informed consent for the use of this data in machine learning and were paid at market rates.

Each interpreter contributed a certain number of hours of work to the task, but there was wide variation in the speed of the interpreters---both in terms of preparation time, and also literally the speed of their signing. For the faster interpreters (\#1 and \#3), we went through a round of revisions where the first author (a non-native but proficient signer) quality-checked all the content and sent feedback to DPAN for rerecordings---minor performance errors like unintentionally omitting details, not high-level changes to the translations, which were of course up to the discretion of the expert interpreters.\footnote{This is another way in which high-quality sign language translation is more difficult than translation for text: the translations are performed in real-time (often based on nonstandardized notes) and cannot be edited later. If there is an error (and if we care about discourse-level tasks, which we do here), the entire video must be rerecorded, which might introduce new errors in the process. This difficulty---which Deaf signers experience when producing new professional-quality content, not just translations---motivates more machine learning research on seamless editing for sign language videos, which is an understudied problem.} Given time constraints, for the other interpreters (\#2 and \#4) we decided to target fewer hours of translated content and do fewer revisions. This means that signer \#1 and \#3 have the most complete and thoroughly revised content, but the others still meet a high quality bar and offer more variation for evaluation. The first author manually annotated caption alignments for all content while performing quality checks.

\begin{table*}[t]
    \centering
    \small
    \setlength{\tabcolsep}{5.5pt}
    \begin{tabular}{lcccccc}
    \toprule
         & \# & \# & \# & sent. length (chars) & sent. duration (secs) & \# \\
         & signers & discourses & sentences & (0, 10, 50, 90, 100)\% & (0, 10, 50, 90, 100)\% & hours \\
        \midrule
        How2Sign & 11 & 2456 & 35263 & 3, 27, 76, 171, 977 & 0.0, 1.7, 5.3, 13.6, 143.0
        & 79.1 \\
        \midrule
        FLEURS-ASL & 5 & 495 & 1749 & 28, 76, 122, 186, 368 & 1.6, 7.4, 13.3, 25.0, 72.8
        & 7.49 \vspace{1mm} \\
        \it signers \\ 
        \hspace{4mm}\#0 &  1 &  103 &  350 &  35, 71, 122, 190, 328 &  4.4, 8.1, 15.1, 24.4, 48.5 &  1.64\\
        \hspace{4mm}\#1 &  1 &  141 &  508 &  35, 81, 123, 183, 368 &  1.6, 6.9, 12.0, 19.6, 46.0 &  1.84 \\
        \hspace{4mm}\#2 &  1 &  70 &  252 &  52, 75, 124, 190, 328 &  4.4, 12.6, 22.7, 37.0, 72.8 &  1.72 \\
        \hspace{4mm}\#3 &  1 &  141 &  494 &  28, 74, 123, 184, 290  &  2.4, 6.6, 10.3, 15.4, 30.5 &  1.57 \\
        \hspace{4mm}\#4 &  1 & 40 &  145 &  33, 80, 117, 165, 246 &  4.3, 8.6, 15.5, 25.6, 38.5 &  0.72\\
    \bottomrule
    \end{tabular}
    \vspace{2mm}
    \caption{\textbf{FLEURS-ASL summary statistics} and comparisons to How2Sign (all splits included). \# of signers, \# of discourses, \# of sentences, sentence length \& duration (across 0th, 10th, 50th, 90th, and 100th percentiles) measured in characters and seconds respectively, and \# of hours.}
    \label{tab:statistics}
\end{table*}

\subsection{Dataset Statistics}

See Table~\ref{tab:statistics} for the final statistics for the translated content in FLEURS-ASL, broken down per signer. Of the 5 signers, 2 are men and 3 are women. Signers \#0-3 are right-handed, and \#4 is left-handed. There is limited variation in skin tone and age; see discussion in Section~\ref{sec:limitations}. Note the variation in median sentence duration across signers (from 10.3 seconds to 22.7 seconds) despite minimal variation in median sentence length (i.e., signing speed varies across interpreters), and the increased median sentence duration for FLEURS-ASL compared to How2Sign (13.2 seconds vs. 5.3 seconds) due to the different domain (web articles vs. informal instructional narratives). See Appendix~\ref{app:disagg} for more description of the individual signers. The short duration of How2Sign clips (especially in the left tail of the distribution) can make them difficult to understand for sentence-level translation in the absence of context~\citep{reconsideringsentence}.

\subsection{Benchmark Tasks}

See Figure~\ref{fig:splits} for a depiction of the final dataset splits. We provide splits for three settings: zero-shot, signer-independent finetuning (i.e., there is no signer overlap between the train and test sets), and signer-dependent finetuning (i.e., all signers overlap).\footnote{We ensure that the signer-dependent splits have disjoint train and test sets in light of the content overlap for signers \#0 and \#1+\#2.} We expect that the signer-dependent splits will be used less than the others, but they may be useful for studying the effects of per-signer personalization.

FLEURS-ASL supports many tasks by virtue of being grafted onto FLORES/FLEURS/Belebele, but we expect that the most common will be sentence- and discourse-level translation from ASL to English. We recommend scoring sentence-level translation with both BLEU~\citep{papineni-etal-2002-bleu} (with sacreBLEU~\citep{post-2018-call} with \texttt{intl} tokenization) and BLEURT~\citep{sellam2020bleurt} (with BLEURT-20~\citep{pu2021learning}). Untimed discourse-level translation can only be scored with BLEU because BLEURT is not intended for long outputs. We score timed translation using Timed BLEU~\citep{cherry21_interspeech} (and by extension, BLEURT), where we temporally interpolate at the granularity of characters rather than tokens for convenience.

\section{Baselines}
\label{sec:baselines}

We provide three sets of baselines. First, we give human baselines for sentence- and discourse-level translation. Second, we train new versions of the YouTube-ASL baselines that can support many FLEURS-ASL sign$\rightarrow$text tasks using a unified multitask mixture, integrating prior text history, timestamps, and a longer context window (rather than just caption-level translation). And third, we evaluate current multimodal frontier models and show that they have virtually no understanding of ASL.

\subsection{Human Baseline}
\label{sec:human-baseline}

We recruited a native Deaf ASL signer with professional training/experience in sign language education through DPAN to perform the human baseline. The annotator was informed of the purpose of their translations (to be released for use in ML research, to understand how well models perform at ASL$\rightarrow$English translation vs. humans) and paid at market rates.

We sampled 8 articles uniformly at random from the signer-dependent test set for each signer. Within each article, we sampled one sentence. For each article, the annotator was tasked with translating that sampled sentence's ASL clip back to English, and then (once finished) was presented with the entire article's ASL video and asked to translate it to English in its entirety. This gives us human baselines for sentence- and discourse-level translation across 40 instances each, where like~\citet{reconsideringsentence} the translations for each setting are performed by the same person (to reduce irrelevant variation) without contaminating the sentence-level translations with discourse context. See Appendix~\ref{app:human-baseline-instructions} for the task instructions.

The scores are 13.0 BLEU (64.6 BLEURT) for sentence-level translation and 13.5 BLEU for discourse-level translation. Like in~\citet{reconsideringsentence}, these metrics (in particular BLEU) may sound low, but qualitatively the human translations look good. The discrepancies with the references come from a variety of sources (natural changes due to backtranslation through a sign language, errors in one of the translation directions) but the most interesting come from visual ambiguity in the sign language translation, such as when ``6,500'' was perceived as ``2,500'' because the 6 was angled such that the ring and middle finger were aligned, making it look like only 2 fingers were raised without close inspection. See Appendix~\ref{app:human-baseline} for the full set of human translations.

\begin{figure*}
    \centering
    \includegraphics[scale=0.82]{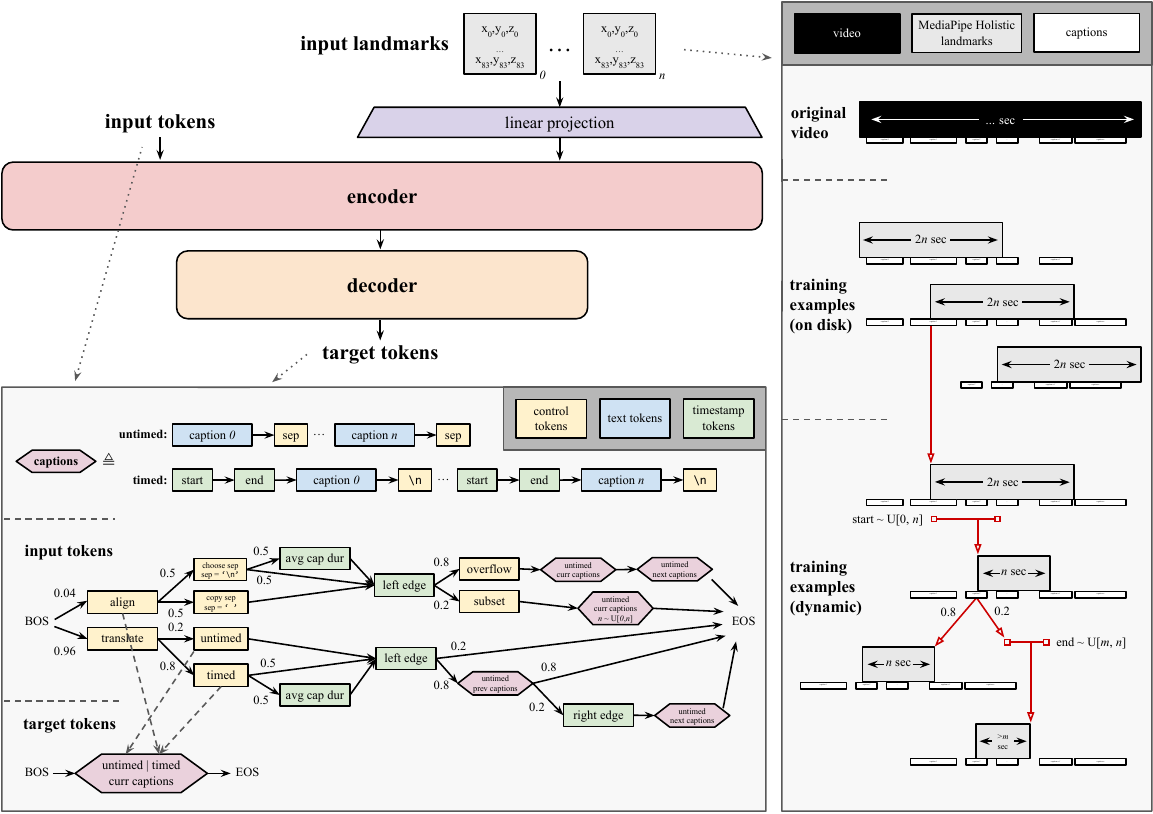} 
    \caption{
    \small
        \textbf{Unified multitask document-level sign to text training.}\vspace{2mm} \newline
        \textbf{\textit{Top left:} Model architecture.} Input text tokens (token embeddings) and up to 512 frames of half-frame-rate linearly projected MediaPipe Holistic landmarks (subset of 85 3D points) are the inputs to T5v1.1-Base (a pretrained encoder-decoder Transformer), finetuned on YouTube-ASL as described below. \vspace{2mm} \newline
        \textbf{\textit{Right:} Preprocessing steps to sample training clips from long videos.} In order to sample training clips uniformly in proportion to video duration, we perform the following steps. (This is only necessary because SeqIO~\citep{roberts2022t5x} does not support sampling training examples in proportion to scalar values attached to them, and because we want to support multi-epoch training with random crops without duplicating the underlying data many times.) We chunk arbitrary length captioned training videos into training examples on disk of $2n$ seconds, preprocess them with MediaPipe Holistic, and carry along the timed caption track, including the previous and next $n$ seconds of captions. (Here, $n$ is 34 seconds so that 15 Hz input fits in 512 tokens.) For each training example, the start position of the clip is sampled from the first half of the chunk uniformly at random, and with probability 0.2 the duration of the clip is truncated to between $m$ and $n$ seconds. (Here, $m$ is 17 seconds). We modify this scheme slightly at the beginning and end of videos; by default, the first and last $n$ seconds of the video would be extremely undersampled because they do not overlap with other examples. We make the chunk duration $\frac{3}{2}n$ seconds rather than $2n$ and sample the start position $\sim max(0, U[-\frac{n}{2}, n])$ for the first chunk and $\sim min(\frac{n}{2}, U[0, \frac{3n}{2}])$ for the last chunk. For videos whose duration doesn't divide evenly by $2n$ seconds, we overlap the final two chunks so they are between $\frac{3n}{2}$ and $2n$ seconds long. \vspace{2mm} \newline
        \textbf{\textit{Bottom left:} Control token format and mixture.} For implementation convenience, we instantiate control tokens as regular text. In order to save compute, task mixture weights are chosen by intuition without hyperparameter sweeps; we expect that optimal weights would change depending on dataset size and application priorities anyway.\vspace{1mm}\newline
        Captions are represented as text spans delimited by separator tokens (either \texttt{` '} or \texttt{`$\backslash$n'}), optionally with start and end timestamps prepended. There are three sets of captions: the ``curr'' captions fully contained within the input clip, the ``prev'' captions starting in the $n$ seconds prior to the input clip, and the ``next'' captions ending in the $n$ seconds after it. There are two main branches of input tokens: caption alignment (with probability 0.04) and translation (0.96). \vspace{1mm}\newline
        The caption alignment branch supports two modes, ``choose sep'' (0.5) where the input specifies caption breaks, and ``copy sep'' (0.5), where the model predicts them. In the latter case, the average caption duration across the source video can be provided as conditioning (0.5) as a form of controllability. Then the ``left edge'' caption timestamp is provided, i.e. the end timestamp of any caption that crosses the left boundary of the landmark input, so that the model knows where to start predicting outputs. Finally, two more modes for either ``overflow'' (0.8), where the provided captions to align may spill past the right boundary of the clip, and ``subset'' (0.2), where the captions are a random prefix of those covering the clip, in order to support whole-video and human-in-the-loop caption alignment respectively. The target tokens are unconditionally timed, newline-separated captions.\vspace{1mm}\newline
        The translation branch supports untimed (0.2) and timed (0.8) translation (determining whether the target captions include timestamp tokens), with the latter also allowing duration conditioning (0.5). Either no captions (0.2), previous captions (0.64), or previous and next captions (0.16) are provided as context, to support isolated translation, blockwise autoregressive translation, and infilling.
    }
    \label{fig:control_tokens}
\end{figure*}

\subsection{Unified Sign to Text Modeling}
\label{sec:method}

\begin{table*}[t]
    \centering
    \small
    \begin{tabular}{l|c|cc}
    \toprule
         & \bf How2Sign & \multicolumn{2}{c}{\bf FLEURS-ASL} \\
         & sentence-level & sentence-level & context+sentence-level \\
        \midrule
        Caption-level & 4.0 (34.4) / - / 12.1 (45.4) & 2.9 (33.6) / 4.8 (38.6) / 5.4 (40.3) & - \\
        Ours & 5.1 (36.9) / - / 14.4 (46.3) & 3.7 (37.2) / 4.9 (39.7) / 6.3 (42.2) & 3.9 (38.3) / 5.3 (41.4) / 6.8 (43.4) \\
        \midrule
        Human & $\sim$ 19.8 (56.6)~\citep{reconsideringsentence} & $\sim$ 13.0 (64.6) & - \\
    \bottomrule
    \end{tabular}
    \vspace{2mm}
    \caption{\textbf{Baseline results for sentence-level translation from ASL to English (zero-shot / signer-independent finetuning / signer-dependent finetuning)}, measured in BLEU (BLEURT in parentheses). We also report scores on How2Sign as a point of reference. Note that the human baseline scores are not perfectly comparable for either benchmark because they are for a subset of the data that is not necessarily representative due to the sampling method and small size.}
    \label{tab:untimed}
\end{table*}

\subsubsection{Method}
\label{sec:setup}

Inspired by Whisper~\citep{whisper} and heeding the call of~\citet{reconsideringsentence} to move beyond sentence-level sign language translation, we present a new unified modeling approach to flexibly support many FLEURS-ASL tasks (and more) in one model. We integrate timestamp tokens and previous text tokens with a 34-second context window trained on random clips from captioned videos. This approach fits into the typical encoder-decoder framework without architectural modifications, only changes to data preparation. See Figure~\ref{fig:control_tokens} for a complete description of this procedure. The key point is that, in contrast to prior work in sign language, our training examples contain caption \textit{tracks} as supervision rather than individual captions, and the timing of these tracks is used to adapt to random video clip boundaries, rather than determine the boundaries of those clips. This disregard for preordained alignments in our clip sampling enables us to handle arbitrary clipping at inference time: in particular, we can perform transformations of long videos in autoregressive chunks, where the next chunk starts after the previous chunk's predicted timestamps.\footnote{Specifically: in addition to providing caption history in the input text tokens, we maintain the first 4 seconds and last 10 seconds of the 34-second context window as regions where predicted captions should not start or stop; if a caption is predicted in that range, we shift the window for autoregression to ensure that the model has more context to perform its translation, and to tolerate misalignment. This heuristic was determined interactively by observing when the autoregression got ``stuck'' on certain videos (not from the test set, which would cause leakage).}

\subsubsection{Experiments}
\label{subsec:evaluation}

We build off YouTube-ASL~\citep{youtubeasl}'s baseline, which linearly projects 85 3D points from MediaPipe Holistic~\citep{lugaresi2019mediapipe, mediapipeholistic} skeletons into the encoder of T5~\citep{t5} at half frame rate, and predicts text tokens from the decoder (with greedy sampling with beam size 5). We compare to a fresh baseline for sentence-level translation so that results are meaningful in light of dataset churn, and report results on context+sentence-level translation (sentence-level translation given prior text history) and discourse-level translation as well. We train each model on 64 TPUv3s for 200k steps with batch size 128 and Adafactor~\citep{shazeer2018adafactor} learning rate 0.001 using the T5X framework~\citep{roberts2022t5x} (10k steps = $\sim$1 hour), with validation BLEU as the checkpoint selection criterion. We finetune on How2Sign for at most 10k steps and FLEURS-ASL for at most 100 steps. We first train the model on caption-level data, as in YouTube-ASL, for 100k steps, then mix in the multitask mixture with weight 0.5.\footnote{This could be seen as a naive kind of context window extension training~\citep{chen2023extendingcontextwindowlarge,peng2023yarnefficientcontextwindow}.} See Tables~\ref{tab:untimed},~\ref{tab:discourse}, and~\ref{tab:alignment} for quantitative results; see Appendix~\ref{app:disagg} for expanded experimental results disaggregated for each individual signer, following~\citet{reconsideringsentence}; and see Table~\ref{tab:qual-examples} for qualitative examples.

The highlights are that our method performs at least as well as the caption-level baseline at sentence-level translation (3.7 vs. 2.9 BLEU, 37.2 vs. 33.6 BLEURT) while being able to tackle additional tasks. The models improve with signer-independent and then signer-dependent finetuning (but much less than the gap on How2Sign), as well as with additional context.
Results are still far below human performance of 13.0 BLEU (64.6 BLEURT). This approach also shows poor performance on caption alignment---far below a strong model-free baseline where caption duration is scaled in proportion to caption length---because the model can decohere from the original input (especially when captions are long relative to the context window) as the model autoregresses. This is not an issue for translation because it does not have to follow along with a given input track.

\begin{table}[t]
    \centering
    \small
    \begin{tabular}{l|cc}
    \toprule
         & \multicolumn{2}{c}{\bf discourse-level} \\
         & untimed & timed \\
        \midrule
        Ours & 4.0 (-) & 3.1 (31.7) \\
        \midrule
        Human & 13.5 (-) & - \\
    \bottomrule
    \end{tabular}
    \vspace{2mm}
    \caption{\textbf{Baseline results for discourse-level translation from ASL to English (untimed vs. timed)}. The untimed column is measured in BLEU, because BLEURT is not intended for long outputs, whereas the timed column is measured in timed BLEU (timed BLEURT in parentheses)~\citep{cherry21_interspeech}. We perform the temporal interpolation for timed BLEU/BLEURT at the character level.}
    \label{tab:discourse}
\end{table}

\begin{table}[t]
    \centering
    \begin{tabular}{lccc}
    \toprule
         & frame-acc \\
        \midrule
        Length scaling & 88.5\% \\
        Ours & 24.3\% \\
    \bottomrule
    \end{tabular}
    \caption{\textbf{Baseline results for ASL-English caption alignment}, measured in frame accuracy as in~\citep{bobsl,bull2021aligning}. ``Length scaling'' refers to a model-free baseline where the caption boundaries are linearly interpolated along the video duration according to the length (in characters) of the corresponding text. Our model often decoheres from the input track while autoregressing and therefore drastically underperforms the length rescaling baseline.}
    \label{tab:alignment}
\end{table}

\subsection{Frontier Models}
\label{subsec:frontier}

Finally, we evaluate 3 multimodal frontier models---Claude 3 Opus~\citep{claude3}, GPT-4o~\citep{gpt4o}, and Gemini 1.5 Pro~\citep{gemini1p5}---on the same FLEURS-ASL sample we used for the human baseline. While Claude 3 Opus only claims support for image inputs (but accepts up to 20 image inputs), GPT-4o and Gemini 1.5 Pro support video, at least at low frame rates. We prompt the model to ``Translate the following video from American Sign Language to English. Give only the translation without any extra explanations.'' and upload the given video at 1 Hz, sampling with default parameters. See quantitative results in Table~\ref{tab:frontier} and qualitative results in Table~\ref{tab:qual-examples}. None of the models show any appreciable sign language understanding. Claude 3 sometimes acknowledges that it cannot understand the videos and chooses not to produce a translation.

Overselling technological progress is all too common in the sign language space and is a \href{https://www.theatlantic.com/technology/archive/2017/11/why-sign-language-gloves-dont-help-deaf-people/545441/}{point of frustration} for the Deaf community. In the brief time that these models have been out, we have personally witnessed multiple independent instances of users testing sign language examples and then declaring that the models must understand sign language, when the examples are either trivial/given away by context or the translations are completely incorrect. We hope that FLEURS-ASL will become a standard addition to frontier model evaluation suites to increase the visibility of sign language tasks and calibrate expectations for them.

\begin{table}[t]
    \centering
    \begin{tabular}{lcc}
    \toprule
         & BLEU & BLEURT \\
        \midrule
        GPT-4o & 0.1 & 32.2 \\
        Claude 3 Opus & 0.4 & 32.0 \\
        Gemini 1.5 Pro & 0.2 & 20.6 \\
        \midrule
        Human & 13.0 & 64.6 \\
    \bottomrule
    \end{tabular}
    \vspace{2mm}
    \caption{\textbf{Performance of frontier multimodal models on sentence-level translation from ASL to English.} None of the models demonstrate any meaningful understanding of ASL; the variation in scores reflects artifacts of output length.}
    \label{tab:frontier}
\end{table}

\begin{table*}[t]
    \centering
    \begin{tabular}{clp{9.2cm}}
    \toprule
        \multirow{4}{*}{(1)} & \bf Reference & During the 1980s he worked on shows such as Taxi, Cheers, and The Tracy Ullman Show. \\
        & Gemini 1.5 Pro & I am learning sign language. \\
        & Ours (zero-shot) & In the 1980s, she worked in theaters like taxesi, cheesy, and tracy. \\
        & Ours (signer-indep ft) & During the 1980s, he worked in theaters such as Axios, Chruas, and the Tracy Ultra Show. \\
        & Ours (signer-dep ft) & During the 1980s, he worked in theaters like taxesi, Cheruns, and the Tracy Ullman Show. \\
        & Human & During the 1980s, they worked in shows like Taxi, Cheers, and The Tracy Ullman show. \\
        \midrule 
        \multirow{4}{*}{(2)} & \bf Reference & The rise of new technologies allows us to see and investigate brain structures and processes never seen before. \\
        & Gemini 1.5 Pro & You must think!  Use your brain! \\
        & Ours (zero-shot) & There is a new technique to detect brains and vision.\\
        & Ours (signer-indep ft) & The increase in new technologies allows a better understanding of the brain and its visual properties. \\
        & Ours (signer-dep ft) & The increase in new technology allows the researchers to recognize brain athletics and brain memory. \\
        & Human & The rise of new technology allows for investigation of brain structure and processes never seen before. \\
        \midrule
        \multirow{4}{*}{(3)} & \bf Reference & The Articles required unanimous consent from all the states before they could be amended and states took the central government so lightly that their representatives were often absent. \\
        & Gemini 1.5 Pro & It is September.  It’s time to go back to school! \\
        & Ours (zero-shot) & The law requires all states to agree on a standard and that it is a legal requirement. \\
        & Ours (signer-indep ft) & The law requires that all states must agree on the same legal regulations. \\
        & Ours (signer-dep ft) & The law requires all states to agree on the same regulations that are not legal. \\
        & Human & The article mandated all states to create a uniform agreement as an amendment to the law for revisions. The states look at the process as minor and many of the state representatives did not make an effort to attend those meetings. \\
    \bottomrule
    \end{tabular}
    \vspace{2mm}
    \caption{\textbf{Qualitative examples from our sentence-level baselines for ASL to English translation on FLEURS-ASL.} Examples selected randomly without cherrypicking. In (2), see that Gemini 1.5 Pro does translate ``brain'', but this is because the sign is iconic (pointing to the head). See Appendix~\ref{app:human-baseline} for the complete set of examples from the human baseline.}
    \label{tab:qual-examples}
\end{table*}

\section{Conclusion}
\label{sec:conclusion}
In this paper, we presented FLEURS-ASL, an extension of the standard FLORES/FLEURS translation benchmarks to their first sign language, American Sign Language. FLEURS-ASL distinguishes itself from prior sign language benchmarks not just with its flexibility (supporting multiple tasks and translation directions by virtue of extending a larger benchmark) but also its domain-specific quality considerations (use of Certified Deaf Interpreters translating English text with substantial preparation time, rather than semi-live interpretation following a video). We provided human baselines, domain-specific model baselines (using a new unified multitask approach that operates on random clips), and multimodal frontier model baselines; these collectively show that there is still a long way to go to reach usable performance on ASL to English translation. We hope that FLEURS-ASL will encourage more focus on this task---framed in a way that deemphasizes strictly aligned captions and therefore admits use of noisier datasets---and serve as scaffolding to extend FLORES/FLEURS to even more of the world's sign languages.

\section{Limitations}
\label{sec:limitations}

While FLEURS-ASL improves upon the quality and flexibility of prior evaluations for sign language understanding, it still has a number of limitations, both in language and vision aspects.

In terms of language: The FLORES~\citep{flores2} sentences were originally drawn from Wikipedia and therefore have a limited domain, favoring third-person over first- and second-person (in monologues only; no dialogues) and lacking Deaf-centric vocabulary and topics. This content was translated from English into ASL, which means that (despite efforts to maintain a high quality bar) the content will surely differ linguistically in some respects from spontaneously uttered ASL. As described in Section~\ref{sec:data-collection}, in order to support sentence-level tasks and caption alignment, we imposed restrictions on reordering of content across sentences that limit discourse structure, and the translation guidelines were tailored primarily to elicit data suitable for evaluating sign language understanding rather than generation.\footnote{Different generation domains demand different styles, which are out of scope here. However, given that the content was translated from English into ASL, it is in some sense actually a closer match for generation than understanding.} FLEURS-ASL currently lacks finer-grained annotations than sentence alignments; we hope that the community could contribute these over time if useful. And of course, FLEURS-ASL only covers American Sign Language, one of the \href{https://www.un.org/en/observances/sign-languages-day}{>300 sign languages} used worldwide.

In terms of vision: The FLEURS-ASL translations are performed in professional recording environments with clean backgrounds, which means that issues relating to camera quality/perspective or environment complexity are not captured by the benchmark. Our interpreters do not cover all phenotypic variation (e.g., skin tone), so the benchmark cannot guarantee fairness across demographics; this is impossible in a set of only 5 signers anyway. We circumvent this problem somewhat in our baselines by using MediaPipe Holistic, which has its own fairness evaluations in its model cards~\citep{mediapipemodelcards}, but the benchmark is intended to be method-agnostic. Many variations in physical appearance correspond with sociolects (e.g., race, gender, and age~\citep{sociolinguistics}), so this is not strictly a vision problem. Note that modeling poses of the kind produced by MediaPipe Holistic is also a fundamentally flawed approach in general because it does not capture semantically relevant features like the tongue (and from the perspective of fairness, cannot handle people with atypically numbered or formed fingers and hands).

Additionally, in terms of truly zero-shot evaluation: Because FLORES/FLEURS is seeded from Wikipedia and widely used for finetuning or evaluation, the text/speech references for FLEURS-ASL may be contaminated in pretraining data. The FLEURS-ASL interpreters may also appear in public data, such as on YouTube, due to the limited pool of suitably qualified candidiates.

\section*{Ethics Statement}

The interpreters who translated FLEURS-ASL were paid at market rates and consented to their video translations being published for use in machine learning research by the broader community. While facial expressions are essential to sign language grammar and therefore must be included in the dataset, we ask that you blur the interpreters' faces when including examples in publications (as we do in this paper). You should not attempt to reidentify the interpreters or use their likenesses to generate and publish other content (deepfakes).

\section*{Acknowledgements}
We would like to thank Sean Forbes, Sam Sepah, and Manfred Georg for contributions to the data collection logistics; Thad Starner, Caroline Pantofaru, Hadar Shemtov, and many others for institutional support; David Uthus and the T5X team for general help with infrastructure; Chris Dyer and Caroline Pantofaru for giving feedback on drafts of this paper; and of course the FLEURS-ASL interpreters and annotators for their participation.

\bibliography{custom}
\bibliographystyle{acl_natbib}

\newpage
\appendix

\section{Translation Guidance}
\label{app:translation-guidance}

While sign language interpreting and translation is a mature field with comprehensive training and certification processes, the desiderata for a machine translation benchmark are somewhat unique compared to the variety of contexts in which interpreting and translation are typically performed. We gave the following guidance to the interpreters to consistently elicit the kind of content we were looking for (not necessarily verbatim; some guidance was provided live during onboarding or as feedback).

\begin{itemize}
    \item This is a challenging set of content to translate, with relatively short narratives about many different topics. We expect that you will spend 5-6 hours preparing the translations for every 1 hour of recorded content to ensure a high quality standard.
    \item Try to sign as naturally as possible. We are looking for content that looks like it was originally signed in ASL rather than translated, as much as possible. For example, this means that you don't need to add explanations of concepts or entities that aren't present in the source content, like you might for a translation of a public service announcement. You should sign as you would in a professional environment to another signer with similar background to you.
    \item Implied by this, do not leave intentional pauses between sentences.
    \item However, you should not reorder content between sentences. i.e., You should be able to take the original sentences and align them to your translation as a caption track, and the captions should be a reasonable translation of the signed content.
    \item If there is any ambiguity in the English sentence when you're translating into ASL, act as if the information was already known from context. For example, if the sentence starts with ``he'', you should not fingerspell ``he'' to clarify the gender, but instead just point and assume that gender is known from context.
    \item If there is anything that needs to be specified when translating into ASL, pick one interpretation and stick with it throughout the narrative. For example, if the sentence is describing a “crash”, without specifying what kind of vehicle crashed (car, plane, etc.), pick the most reasonable interpretation given the context and go with it.
\end{itemize}

\section{Human Baseline Instructions}
\label{app:human-baseline-instructions}

Your task today will be to perform ASL$\rightarrow$English translation tasks on a sample of 40 videos we provide to you. We will use/release these translations so that we and the broader AI research community can compare how well AI translates ASL vs. a human (and get a better understanding of its limitations).

\textbf{You’re allowed to watch the video multiple times, slow down the video, etc. to help you.} But don’t agonize too much over it.

\textbf{You must perform the tasks in a particular order, in order to ensure the validity of the results.} Specifically, we will ask you to:

\begin{itemize}
    \item[(a)] Look at a clip containing 1 sentence, and translate that clip.
    \item[(b)] Look at a full video containing that sentence plus others around it, and translate the entire video given the additional context.
\end{itemize}

First, we’ll go through an example.

a) Watch the following clip. The task is to translate the clip. My translation of the clip is:

\textit{If you stand at the edge of the water, you’ll look down to the bottom and see pebbles and gunk.}

b) Watch the following clip. The task is to translate the entire video. (Note that in light of additional context, you may refine your translation from step a for step b.) My translation of the video:

\textit{It will behave similar to water. It’s transparent like water. If you stand at the shore and look down, you’ll see through it to the pebbles and gunk on the bottom. Stofan adds, “As far as we know, only one planet is more active than Titan: the planet Earth.”}

\section{Disaggregated Analysis}
\label{app:disagg}

First, some subjective observations about the different signers' styles:

Signers \#0, \#3, and \#4 tend to sign closer to the original English syntax when there is no strong reason not to, whereas signers \#1 and \#2 more freely restructure the sentences. Some individual points of distinction:

\begin{itemize}
    \item Signer \#0 performs a swish to separate individual words in a fingerspelled phrase, which is uncommon.
    \item Signer \#1 tends to use more rare signs (such as cities/countries) and vanguard language practices (such as using the endogenous BSL sign for England rather than the traditional ASL sign).
    \item Signer \#2 signs the slowest.
    \item Signer \#3 signs the fastest.
    \item Signer \#4 signs primarily left-handed, but is more ambidextrous than the others.
\end{itemize}

Second, see Table~\ref{tab:disagg} for disaggregated sentence-level ASL to English translation results. As expected due to our more stringent set of requirements for recruiting interpreters, the human baseline scores (especially BLEURT) in FLEURS-ASL are more consistent across signers than the scores on How2Sign provided by~\citet{reconsideringsentence}. The BLEURT scores are also higher overall; we speculate that this is due factors such as the source sentences being longer, the translations into ASL being higher quality/more faithful, and the reference domain being more neutral. The variation in scores that remains in FLEURS-ASL does not seem to reflect underlying translation quality, but just the depth of the translation (and e.g., amount of fingerspelling).

\begin{table}[H]
    \centering
    \small
    \begin{tabular}{l|cc}
    \toprule
         & sentence-level \\
        \midrule
        Caption-level & \\
        \hspace{4mm}\#0 & 3.6 (34.8) \\
        \hspace{4mm}\#1 & 2.0 (31.8) \\
        \hspace{4mm}\#2 & 3.6 (35.7) \\
        \hspace{4mm}\#3 & 4.2 (36.2) \\
        \hspace{4mm}\#4 & 3.9 (33.8) \\
        \midrule
        Ours & \\
        \hspace{4mm}\#0 & 5.1 (39.1) \\
        \hspace{4mm}\#1 & 2.4 (35.1) \\
        \hspace{4mm}\#2 & 4.3 (38.8) \\
        \hspace{4mm}\#3 & 5.3 (39.5) \\
        \hspace{4mm}\#4 & 4.7 (39.1) \\
        \midrule
        Human\vspace{1mm} & \\
        \hspace{4mm}\#0 & 12.1 (63.4) \\
        \hspace{4mm}\#1 & 9.0 (60.6) \\
        \hspace{4mm}\#2 & 6.5 (63.1) \\
        \hspace{4mm}\#3 & 16.1 (68.5) \\
        \hspace{4mm}\#4 & 21.7 (67.6) \\
    \bottomrule
    \end{tabular}
    \vspace{2mm}
    \caption{\textbf{Results for zero-shot sentence-level translation from ASL to English disaggregated by signer}, measured in BLEU (BLEURT in parentheses).}
    \label{tab:disagg}
\end{table}

\section{Complete Human Baseline}
\label{app:human-baseline}

See Table~\ref{tab:human0} for the complete set of FLEURS-ASL human baseline translations, also available at \datalink.

\clearpage
\begin{table*}[]
    \caption{\textbf{Complete set of FLEURS-ASL human baseline translations}. We provide one sentence-level translation (performed in isolation) and then the complete discourse-level translation for each document sequentially.}
    \centering
    \fontsize{8.4}{10}\selectfont
    \begin{tabular}{cp{7cm}p{7cm}}
        \toprule
        \textbf{signer} & \textbf{reference} & \textbf{human prediction} \\
\bottomrule \vspace{-2.5mm} \\
0 & As NSA, he assisted Carter in diplomatically handling world affairs, such as the Camp David Accords, 1978; normalizing US–China relations thought the late 1970s; the Iranian Revolution, which led to the Iran hostage crisis, 1979; and the Soviet invasion in Afghanistan, 1979. & NSL’s work role supported Carter’s cabinet to handle world events. For example, Camp David Accords of 1978, National Customs of USA and China relations in late 1970s, the Iraqi revolution that resulted in the Iraqi hostage crisis in 1979, and the Soviet takeover of Afghanistan in 1979. \\
\midrule
& Throughout 1960s, Brzezinski worked for John F. Kennedy as his advisor and then the Lyndon B. Johnson administration. During the 1976 selections he advised Carter on foreign policy, then served as National Security Advisor (NSA) from 1977 to 1981, succeeding Henry Kissinger. As NSA, he assisted Carter in diplomatically handling world affairs, such as the Camp David Accords, 1978; normalizing US–China relations thought the late 1970s; the Iranian Revolution, which led to the Iran hostage crisis, 1979; and the Soviet invasion in Afghanistan, 1979. & In the 1960s, a person named Burzezinski worked for John K. F. Kennedy in his role as an advisor. His role as advisor continued with the next president, Lyndon B Johnson. During the 1976 selection, he advised Carter on foreign policy and then served a role in NSA, National Security Advisory from 1977 to 1981, and then eventually handed his role over to Henry Kissinger.  His role in NSA was to support Carter in policies handling world events.  For example, Camp David Accords of 1978, National Customs between USA and China relations in late 1970s, Iraqi war that resulted in the Iraqi hostage crisis in 1979, and the Soviet takeover of Afghanistan in 1979. \\
\bottomrule \vspace{-2.5mm} \\
0 & Late on Sunday, the United States President Donald Trump, in a statement delivered via the press secretary, announced US troops would be leaving Syria. & All day into the night, late at night Sunday, the United States president Donald Trump sent a note to the press secretary to announce that the US troops will go overseas to Syria. \\
\midrule
& Late on Sunday, the United States President Donald Trump, in a statement delivered via the press secretary, announced US troops would be leaving Syria. The announcement was made after Trump had a phone conversation with Turkish President Recep Tayyip Erdoğan. Turkey would also take over guarding captured ISIS fighters which, the statement said, European nations have refused to repatriate. & All day into the night, late Sunday, president Donald Trump sent a note to the press secretary to announce that US troops will leave Syria. After the announcement, Trump met with Turkey’s president Recep Tayyip Erdogăn. Turkey will also take responsibility in watching over the ISIS militant captives. There was a note that European countries will not allow these captives back into their countries. \\
\bottomrule \vspace{-2.5mm} \\
0 & Mrs. Kirchner announced her intention to run for president at the Argentine Theatre, the same location she used to start her 2005 campaign for the Senate as member of the Buenos Aires province delegation. & They announced their intention of their campaign for president at a guitar theater.  Understand that the guitar theater is the same area as the 2005 campaign as a senator.  Same member started the regional president demonstration. \\
\midrule
& Current senator and Argentine First Lady Cristina Fernandez de Kirchner announced her presidential candidacy yesterday evening in La Plata, a city 50 kilometers (31 miles) away from Buenos Aires. Mrs. Kirchner announced her intention to run for president at the Argentine Theatre, the same location she used to start her 2005 campaign for the Senate as member of the Buenos Aires province delegation. & The current senate is also the Argentine president’s wife, Cristina Fernandez de Kirchner.  She announced presidential candidacy last night at La Plata about 50 km, or 31 miles, from Buenos Aires.  She announced her intention of running for presidency at a theater.  Understand the theater is the same place where she announced she was running for candidacy for senate in 2005 as a member of Buenos Aires Local President Demonstration. \\
\bottomrule \vspace{-2.5mm} \\
0 & The National Hurricane Center (NHC) says that at this point Jerry poses no threat to land. & NHC, National Hurricane Center, commented that today Jerry was not severe in affecting the land. \\
\midrule
& The tenth named storm of the Atlantic Hurricane season, Subtropical Storm Jerry, formed in the Atlantic Ocean today. The National Hurricane Center (NHC) says that at this point Jerry poses no threat to land. & The tenth storm during the Atlantic Hurricane season, named Jerry, was first named as a subtropical storm in the Atlantic ocean.  NHC, National Hurricane Center, commented today that Jerry was not severe. \\
\bottomrule \vspace{-2.5mm} \\
    \end{tabular}
    \label{tab:human0}
\end{table*}

\clearpage
\begin{table*}[]
    \centering
    \fontsize{8.4}{10}\selectfont
    \begin{tabular}{cp{7cm}p{7cm}}
        \toprule
        \textbf{signer} & \textbf{reference} & \textbf{human prediction} \\
\bottomrule \vspace{-2.5mm} \\
0 & According to the lawsuit, waste from the UN camp was not properly sanitized, causing bacteria to enter the tributary of the Artibonite River, one of Haiti's largest. & The lawsuit follows the United Nations camp and their waste as it is not properly sanitized.  That caused the pollution to enter the Artibonite River, which is one of Haiti's largest river. \\
\midrule
& UN peacekeepers, whom arrived in Haiti after the 2010 earthquake, are being blamed for the spread of the disease which started near the troop's encampment. According to the lawsuit, waste from the UN camp was not properly sanitized, causing bacteria to enter the tributary of the Artibonite River, one of Haiti's largest. Prior to the arrival of troops, Haiti had not encountered problems related to the disease since the 1800s. The Haitian Institute for Justice and Democracy has referenced independent studies that suggest the Nepalese UN peacekeeping battalion unknowingly brought the disease to Haiti. Danielle Lantagne, a UN expert on the disease, stated the outbreak was likely caused by the peacekeepers. & The United Nations Peacekeepers arrived in Haiti and began monitoring after the 2010 earthquake, where they started to blame the spread of disease on the military.  The lawsuit documents that the United Nations camp’s waste is not properly sanitized so it caused the pollution to enter the Artibonite River, one of Haiti's largest rivers.  Before the military arrived, there were no issues with Haiti since the 1800s.  So now once the military arrived, the issue came up.  The Nile Institute Justice Department was involved with studies and suggested that the Nepal United Nations military were unaware that they were bringing diseases to Haiti.  The United Nations disease expert named Danielle Lantagne made a statement that the spread was started by the United Nations Peacekeepers. \\
\bottomrule \vspace{-2.5mm} \\
0 & During the 1980s he worked on shows such as Taxi, Cheers, and The Tracy Ullman Show. & During the 1980s, they worked in shows like Taxi, Cheers, and The Tracy Ullman show. \\
\midrule
& Before The Simpsons Simon had worked on several shows in various positions. During the 1980s he worked on shows such as Taxi, Cheers, and The Tracy Ullman Show. In 1989 he helped create The Simpsons with Brooks and Groening, and was responsible for hiring the show's first writing team. Despite leaving the show in 1993 he kept the title of executive producer, and continued to receive tens of millions of dollars every season in royalties. & Before The Simpsons, Simon worked in several different shows in different roles. During the 1980s, he worked in different shows like Taxi, Cheers and the Tracy Ullman show.  In 1989, he helped implement The Simpsons with Brooks and Groening.  He was responsible for hiring the first writers for the show.  In 1993, even though he was very successful, he maintained his title as Executive Producer and earned tens of millions of dollars in royalty every season. \\
\bottomrule \vspace{-2.5mm} \\
0 & Her other race, the Giant Slalom, saw her finish in tenth in the women's sitting group with a combined run time of 4:41.30, 2:11.60 minutes slower than first place finisher Austrian Claudia Loesch and 1:09.02 minutes slower than the ninth place finisher Gyöngyi Dani of Hungary. & Another person who competed is Giant Slalom who finished tenth place.  The women who were sitting in the group, their total time together was 4 hours, 41 minutes and 30 seconds. In comparison with the first place winner, who was from Amsterdam named Claudia Loesch, they were about 2 hours, 11 minutes, and 60 seconds slower.  9th place was Giöngyi Dani from Hungary, who was 1 hour, 9 minutes, and 2 seconds slower by comparison. \\
\midrule
& Beyond Wednesday's event, Carpanedo competed in two individual races at the Championships. Her first was the Slalom, where she earned a Did Not Finish in her first run. 36 of the 116 competitors had the same result in that race. Her other race, the Giant Slalom, saw her finish in tenth in the women's sitting group with a combined run time of 4:41.30, 2:11.60 minutes slower than first place finisher Austrian Claudia Loesch and 1:09.02 minutes slower than the ninth place finisher Gyöngyi Dani of Hungary. Four skiers in the women's sitting group failed to finish their runs, and 45 of the 117 total skiers in the Giant Slalom failed to rank in the race. & In Wednesday’s event in Carpenedo there were two individuals in a championship tournament.  The first competitor was Slalom who earned the title of ‘Did Not Finish’ in the first run.  Out of 116 competitions, 36 got the same result of ‘Did Not Finish’.  The other competitor named Giant Slalom finished tenth place.  The women in the sitting group's total time competing was 4 hours, 41 minutes and 30 seconds.  In comparison with the first place winner, who was from Amsterdam named Claudia Loesch, they were about 2 hours, 11 minutes and 60 seconds slower.  Also in comparison with the 9th place winner who was Giöngyi Dani from Hungary, they were 1 hour, 9 minutes, and 2 seconds slower.  Four of the skiers in the women sitting group flunked out of the competition.  Out of 117 competitors, 45 failed to move up in the tournament.  Giant Slalom also failed and never moved up in the tournament. \\
\bottomrule \vspace{-2.5mm} \\
0 & The cabbage juice changes color depending on how acidic or basic (alkaline) the chemical is. & Cabbage juice changes colors depending on how acidic it is and how basic, or the alkaline chemical, it is. \\
\midrule
& This is called a chemical's pH. You can make an indicator using red cabbage juice. The cabbage juice changes color depending on how acidic or basic (alkaline) the chemical is. The pH level is indicated by the amount of Hydrogen (the H in pH) ions in the tested chemical. Hydrogen ions are protons that had their electrons stripped off them (since Hydrogen atoms consist of one proton and one electron). & This is called chemical pH.  You can compute pH levels by using red cabbage juice. The juice will change colors depending on the mix of two components, where the first is acidic and the second is basic, and the two put together is the alkaline chemical that influences the color change in cabbage juice.  The pH levels are detected by the amount of hydrogen within the pH ions.  Hydrogen ions are protons but the electrons will take ions away.  The hydrogen ion only has one proton and one electron. \\
\bottomrule \vspace{-2.5mm} \\
    \end{tabular}
    \label{tab:human1}
\end{table*}

\clearpage
\begin{table*}[]
    \centering
    \fontsize{8.4}{10}\selectfont
    \begin{tabular}{cp{7cm}p{7cm}}
        \toprule
        \textbf{signer} & \textbf{reference} & \textbf{human prediction} \\
\bottomrule \vspace{-2.5mm} \\
1 & Tournament top seeds South Africa started on the right note when they had a comfortable 26 - 00 win against 5th seeded Zambia. & In the tournament bracket, South Africa moved forward strong and won 26 to 0 against Zambia, who was 5th. \\
\midrule
& The games kicked off at 10:00am with great weather and apart from mid morning drizzle which quickly cleared up, it was a perfect day for 7's rugby. Tournament top seeds South Africa started on the right note when they had a comfortable 26 - 00 win against 5th seeded Zambia. Looking decidedly rusty in the game against their southern sisters, South Africa however steadily improved as the tournament progressed. Their disciplined defence, ball handling skills and excellent team work made them stand out and it was clear that this was the team to beat. & The game started at 10am and the weather was perfect, even though in the morning it was a bit hazy, but has since then disappeared. The rugby game was a 7-on-7 and they followed a tournament bracket. South Africa progressed nicely in the tournament and won 26-0 against Zambia, who was 5th. In the beginning of the game, it was a bit rough but eventually smoothed over. South Africa, regardless, improved throughout the tournament and became stronger as they progressed. Their defense worked seamlessly together and handled the ball nicely. The team showed solidarity. That’s the team that beat the other. \\
\bottomrule \vspace{-2.5mm} \\
1 & Others are alleged to have been brought up by animals; some are said to have lived in the wild on their own. & Some of them ventured off and joined the animals while others were isolated. \\
\midrule
& One of the most common methods used to illustrate the importance of socialization is to draw upon the few unfortunate cases of children who were, through neglect, misfortune, or wilful abuse, not socialized by adults while they were growing up. Such children are called "feral" or wild. Some feral children have been confined by people (usually their own parents); in some cases this child abandonment was due to the parents' rejection of a child's severe intellectual or physical impairment. Feral children may have experienced severe child abuse or trauma before being abandoned or running away. Others are alleged to have been brought up by animals; some are said to have lived in the wild on their own. When completely brought up by non-human animals, the feral child exhibits behaviors (within physical limits) almost entirely like those of the particular care-animal, such as its fear of or indifference to humans. & There are discussions on the importance of socializing. It always goes back to children who grow up with no exposure to healthy interactions with adults, and this can happen for various reasons, some being neglect, bad luck, or just abuse. When children don’t get social exposure, they become what you call feral, or “wild child”. Isolating their children may be the result of some people being ashamed of these children because of their mental state or their physical appearance. Sometimes it’s their own parents doing this. What happens is they may sometimes escape captive because of trauma from mental abuse, sometimes severe. Sometimes these children escape and end up living with animals, being the only human. There is a trend that these children will mimic the body language and behavior of the animal they’ve interacted with. Sometimes these feral children are afraid of other humans while others do not care to make any connections with humans. \\
\bottomrule \vspace{-2.5mm} \\
1 & His teammate Fernando Alonso was in the lead for most of the race, but ended it right after his pit-stop, probably because a badly tucked right front wheel. & The other team, the same as Fernado Alonso, drove and eventually spun out.  He had to pull over to the pit stop because one of his tires was bent inward.  They had to remove the car from the race. \\
\midrule
& Giancarlo Fisichella lost control of his car and ended the race very soon after the start. His teammate Fernando Alonso was in the lead for most of the race, but ended it right after his pit-stop, probably because a badly tucked right front wheel. Michael Schumacher ended his race not long after Alonso, because of the suspension damage in the numerous battles during the race. & In the car race, Giancauldo Fischla spun out and his car was damaged.  He was out.  The other team, like him, Fernando Alonso, also spun out of control and he had to pull over to the pit stop.  One of the tires was bent inward and they determined that the driver was also out of the race.  Another car, Michael Schumacher, after the other two were already out, was also out because his suspensions were not functioning properly. \\
\bottomrule \vspace{-2.5mm} \\
1 & Today many Sámi work in modern trades. Tourism is an important income in Sápmi, the Sámi area. & Today many Sami work in their world as their job.  Tourism, for example, Sapmi is a cash cow as a form of unification. \\
\midrule
& Reindeer husbandry is an important livelihood among the Sámi and the culture surrounding the trade is important also for many with other professions. Even traditionally, though, not all Sámi have been involved in big scale reindeer husbandry, but lived from fishing, hunting and similar, having reindeer mostly as draft animals. Today many Sámi work in modern trades. Tourism is an important income in Sápmi, the Sámi area. & Caring and nurturing reindeer to ensure growth is important.  Sami people and their culture depend on reindeer. It just so happens that other work fields are also associated with reindeer.  It is not intentional for all Sami people to be involved with reindeer but living from fishing and hunting included the reliance and use of reindeer.  Now in today’s world, many Sami workers in tourism, for example, bring in a lot of money, Sapmi, and they are unified by that too. \\
\bottomrule \vspace{-2.5mm} \\
    \end{tabular}
    \label{tab:human2}
\end{table*}

\clearpage
\begin{table*}[]
    \centering
    \fontsize{8.4}{10}\selectfont
    \begin{tabular}{cp{7cm}p{7cm}}
        \toprule
        \textbf{signer} & \textbf{reference} & \textbf{human prediction} \\
\bottomrule \vspace{-2.5mm} \\
1 & But Prime Minister John Howard has said the act was only to safeguard the facilities of the hospital from being downgraded by the Tasmanian government, in giving an extra AUD\$45 million. & The priest John Howard said the act is for protecting the hospital from worsening.  They cherish it and want to put up 45 million AUD. \\
\midrule
& Prime Minister Stephen Harper has agreed to send the government's 'Clean Air Act' to an all-party committee for review, before its second reading, after Tuesday's 25 minute meeting with NDP leader Jack Layton at the PMO. Layton had asked for changes to the conservatives' environmental bill during the meeting with the PM, asking for a "thorough and complete rewriting" of the Conservative party's environmental bill. Ever since the Federal Government stepped in to take over funding of the Mersey hospital in Devonport, Tasmania, the state government and some federal MPs have criticised this act as a stunt in the prelude to the federal election to be called by November. But Prime Minister John Howard has said the act was only to safeguard the facilities of the hospital from being downgraded by the Tasmanian government, in giving an extra AUD\$45 million. & The minister Stephen Harper agreed he will send his government’s Clean Air Act to all political parties to be looked over.  There was a brief 25-minute meeting last week Tuesday with NDP’s leader, Jaclyn Pmo, and Jacklyn said the act needed to be reviewed and revised because it doesn’t suit their conservative environment.  Understand that the government has been involved with funding for Mersey Hospital in Deveonport, Tasmania.  The government MP oversaw the whole process and came to the conclusion that the NDP intentionally shifted their focus due to elections coming up in November.  Something was not right.  However, the priest John Howard clarified that the act is for protecting the hospital from worsening.  They cherish it and want to send forth 45 million AUD. \\
\bottomrule \vspace{-2.5mm} \\
1 & If you booked your flights and accommodation for 2020 before the postponement was announced, you may have a tricky situation. & Have you already booked a flight and hotel for a trip in 2020?  Before they had announced postponement, maybe you were faced with challenges and how to resolve them. \\
\midrule
& The Paralympics will take place from 24 August to 5 September 2021. Some events will be held in other locations throughout Japan. Tokyo will be the only Asian city to have hosted two summer Olympics, having hosted the games in 1964. If you booked your flights and accommodation for 2020 before the postponement was announced, you may have a tricky situation. Cancellation policies vary, but as of late March most coronavirus-based cancellation policies don't extend to July 2020, when the Olympics had been scheduled. It's expected that most event tickets will cost between ¥2,500 and ¥130,000, with typical tickets costing around ¥7,000. & The Paralympics will be hosted from August 24th to September 5th, 2021 in Japan.  Tokyo is the only Asian city that will have hosted two summer games, the first time being in 1964.  If you already booked a flight and hotel for 2020 before the announcement of the postponement, maybe you were faced with challenges and what to do.  The cancellation policies were all over the place.  Recently in March, they had determined that during the time of COVID affecting cancellations, it doesn’t apply to or affect their schedule announced in July 2020.  It is expected that most tickets will cost 2,500 to 130,000 yen, with the most average cost being 7,000 yen. \\
\bottomrule \vspace{-2.5mm} \\
1 & The rise of new technologies allows us to see and investigate brain structures and processes never seen before. & The rise of new technology allows for investigation of brain structure and processes never seen before. \\
\midrule
& Neurobiological data provide physical evidence for a theoretical approach to the investigation of cognition. Therefore it narrows the research area and makes it much more exact. The correlation between brain pathology and behaviour supports scientists in their research. It has been known for a long time that different types of brain damage, traumas, lesions, and tumours affect behaviour and cause changes in some mental functions. The rise of new technologies allows us to see and investigate brain structures and processes never seen before. This provides us with a lot of information and material to build simulation models which help us to understand processes in our mind. & Neurobiological data provides evidence of theoretical approaches and investigation related to cognition, more specifically for the research area to be exact.  There has been an established connection between brain pathology and behavior that has supported science and their research so far.  It has already been proven that brain damage, trauma, lesions and tumors affect behaviors.  These affect some changes to occur in mental functions.  The rises in new technology allows for investigation of brain structure and processes never seen before. It provides new information and resources to build a model that we can mimic to help us understand the brain processes that are happening. \\
\bottomrule \vspace{-2.5mm} \\
1 & Courier companies are well paid for delivering things quickly. Frequently, time is very important with business documents, merchandise or spare parts for an urgent repair. & Courier companies are paid well for fast delivery. What happens often is these things are time sensitive such as documents related to business, materials and extra parts for quick repair. \\
\midrule
& Courier companies are well paid for delivering things quickly. Frequently, time is very important with business documents, merchandise or spare parts for an urgent repair. On some routes, the larger companies have their own planes, but for other routes and smaller firms there was a problem. If they sent things by air freight, on some routes it may have taken days to get through unloading and customs. The only way to get it through faster was to send it as checked luggage. Airline regulations will not allow them to send luggage without a passenger, which is where you come in. & Courier companies are paid well for fast delivery. Oftentimes, these things being delivered are time sensitive such as business documents, materials and extra materials for quick repair.  Larger companies have the luxury of convenient route options such as flights while others have limited options, like smaller companies with less options.  Problems may arise for them.  When these companies ship via flight, some will have to go through customs and investigations prolonging the process, so they expedite the process by putting these items in their own luggages onto a flight.  However, flight regulations do require that luggages travel with their person and that is what the role of a courier is. \\
\bottomrule \vspace{-2.5mm} \\
    \end{tabular}
    \label{tab:human3}
\end{table*}

\clearpage
\begin{table*}[]
    \centering
    \fontsize{8.4}{10}\selectfont
    \begin{tabular}{cp{7cm}p{7cm}}
        \toprule
        \textbf{signer} & \textbf{reference} & \textbf{human prediction} \\
\bottomrule \vspace{-2.5mm} \\
2 & The Articles required unanimous consent from all the states before they could be amended and states took the central government so lightly that their representatives were often absent. & A legal article requires that all states must create an agreement.  That can be an amendment to the law for revisions. The states look at the law as minor.  Many state representatives come together regularly, but many do not make an effort to attend the meetings. \\
\midrule
& During the Revolutionary War, the thirteen states first formed a weak central government—with the Congress being its only component—under the Articles of Confederation. Congress lacked any power to impose taxes, and, because there was no national executive or judiciary, it relied on state authorities, who were often uncooperative, to enforce all its acts. It also had no authority to override tax laws and tariffs between states. The Articles required unanimous consent from all the states before they could be amended and states took the central government so lightly that their representatives were often absent. & During the Revolutionary War, there were 13 states that came together to create a central government. The government was weak because there was one congress that was established by the Article of Confederation. Congress had no power to enforce taxation on people because there was no Executive branch, where the president falls, and then the other branch which is Judicial. So the government relies on the states’ authority for taxes.  However, the government can try to enforce the law but the people won’t cooperate. The government struggled with oppression in the different states and their tax laws and their tariffs between the states. The article mandated all states to create a uniform agreement as an amendment to the law for revisions. The states look at the process as minor and many of the state representatives did not make an effort to attend those meetings. \\
\bottomrule \vspace{-2.5mm} \\
2 & The largest tournament of the year takes place in December at the polo fields in Las Cañitas. & The largest tournament of the year is in Las Canitas in December. \\
\midrule
& Argentina is well known for having one of the best polo teams and players in the world. The largest tournament of the year takes place in December at the polo fields in Las Cañitas. Smaller tournaments and matches can also be seen here at other times of the year. For news on tournaments and where to buy tickets for polo matches, check Asociacion Argentina de Polo. & In Argentina, they are well known for having one of the best polo teams and players in the world. The largest tournament of the year is in Las Canitas in December. Year round, at any time of the year, there are smaller tournaments that take place. If you want to know more information about the tournaments and where you can buy tickets for the polo games, check out Association of Argentina de Polo. \\
\bottomrule \vspace{-2.5mm} \\
2 & After seeing the horrors and atrocities of war during World War I, nations desired to avoid such a situation again in the future. & During WWI, there were a lot of battles and people were there to witness the violence, cruelty, horror, and disruption.  After the war, countries did their best to avoid any conflicts or bad situations from happening again.  That was resolved. \\
\midrule
& During the 1920s, the prevailing attitudes of most citizens and nations was that of pacifism and isolation. After seeing the horrors and atrocities of war during World War I, nations desired to avoid such a situation again in the future. & Back in the 1920s, during that time most people had a relaxed attitude and were open with each other and had honest dialogues.  It was a peaceful time.  Even at the country level, things were civil and peaceful.  Issues that arose were resolved.  There were unique situations where certain countries were isolated.  Then during WWI, there were tense battles and many people saw the violence, cruelty, horror and disruptions that occurred.  After the war was over, countries did their best to avoid the same situation happening again.  That was resolved. \\
\bottomrule \vspace{-2.5mm} \\
2 & Children are placed in Foster Care for a wide variety of reasons that range from neglect, to abuse, and even to extortion. & Children are in foster care for different reasons like neglect, abuse, sexual abuse, and even for forcing, threatening, bribing or grooming. \\
\midrule
& Children are placed in Foster Care for a wide variety of reasons that range from neglect, to abuse, and even to extortion. No child should ever have to grow up in an environment that is not nurturing, caring, and educational, but they do. We perceive the Foster Care System to be a safety zone for these children. Our foster care system is supposed to provide safe homes, loving caregivers, stable education, and reliable health care. Foster care is supposed to provide all the necessities that were lacking in the home they were previously taken from. & Children are in foster care for different reasons like neglect, physical abuse, sexual abuse, and even for forcing, threatening, bribing or grooming.  A child should not grow up in an environment that is not nurturing or thriving in self esteem or education.  It is unfortunate for those who grow up in that environment.  It is important to observe children are in a safe area when in foster care.  Foster care all over must provide a loving, caring and stable environment, as well as reliable health care.  Recently a child was taken from their home due to their needs not being met.  Foster care is supposed to provide the needs that their original home did not provide. \\
\bottomrule \vspace{-2.5mm} \\
    \end{tabular}
    \label{tab:human4}
\end{table*}

\clearpage
\begin{table*}[]
    \centering
    \fontsize{8.4}{10}\selectfont
    \begin{tabular}{cp{7cm}p{7cm}}
        \toprule
        \textbf{signer} & \textbf{reference} & \textbf{human prediction} \\
\bottomrule \vspace{-2.5mm} \\
2 & One of the most noteworthy recent examples of this was the North Atlantic campaign of WWII. The Americans were trying to move men and materials across the Atlantic Ocean to help Britain. & One important, recent event to recognize is in World War II and their North Atlantic Campaign, North America brought over their men and materials over the Atlantic Ocean with the intention of helping Britain. \\
\midrule
& Using ships to transport goods is by far the most efficient way to move large amounts of people and goods across oceans. The job of navies has traditionally been to ensure that your country maintains the ability to move your people and goods, while at the same time, interfering with your enemy's ability to move his people and goods. One of the most noteworthy recent examples of this was the North Atlantic campaign of WWII. The Americans were trying to move men and materials across the Atlantic Ocean to help Britain. At the same time, the German navy, using mainly U-boats, was trying to stop this traffic. Had the Allies failed, Germany probably would have been able to conquer Britain as it had the rest of Europe. & There is the ocean and then land on both sides. A lot of people brought materials over the ocean via ship as it was the most effective method of transportation.  The navy warships would oversee the process of transporting their people and materials.  Also, when they would see the enemy transporting their people and materials, they would interfere with their transport.  One important event recognized recently is during World War II and their North Atlantic campaign, where they would bring over their men and materials overseas to Britain in order to help them, Germany attempted to interfere with transport using their U-boats.  If the alliance and transportation lines fell apart, that means there would be the possibility of Germany having been able to take over Britain and eventually the rest of Europe. \\
\bottomrule \vspace{-2.5mm} \\
2 & This term derives from ancient familiarity with Bed-bugs, which are insects highly adapted to parasitize humans. & The word bug comes from the word “bed bug”, which is an insect that pierces and burrows under the skin, and becomes a parasite. \\
\midrule
& The term bug is used by entomologists in a formal sense for this group of insects. This term derives from ancient familiarity with Bed-bugs, which are insects highly adapted to parasitize humans. Both Assassin-bugs and Bed-bugs are nidicolous, adapted to living in nest or housing of their host. & People who study insects were looking over a subgroup of insects that included the word ‘bug’.  The word comes from an insect called ‘Bed Bugs’ that would pierce and burrow under the skin, and become a parasite.  There are two different types of these bugs, Assassin Bugs and Bed Bugs, which tend to adapt and remain in the body.  They essentially nest in the body for as long as they can. \\
\bottomrule \vspace{-2.5mm} \\
2 & For over a thousand years the Christian religion had bound European states together despite differences in language and customs. I & Thousands of years in Europe and in their different regions, Christians stayed together regardless of different languages or customs. \\
\midrule
& During this period of European history, the Catholic Church, which had become rich and powerful, came under scrutiny. For over a thousand years the Christian religion had bound European states together despite differences in language and customs. I Its all-pervading power affected everyone from king to commoner. One of the main Christian tenets is that wealth should be used to alleviate suffering and poverty and that the monetary funds of the church are there specifically for that reason. The central authority of the church had been in Rome for over a thousand years and this concentration of power and money led many to question whether this tenet was being met. & In Europe’s historical time period where catholic churches were wealthy and powerful, all eyes were on them.  In the thousands of years in Europe and their different regions, Christians stuck together regardless of their different languages or customs.  Religion was powerful and had influence everywhere such as monarchy and even peasants.  Christians had a belief system and one part says that the wealthy shall help the poor and reduce their suffering.  WIth this in mind, the Church would save their money with the intention of reducing suffering in the poor.  Over the thousands of years, the central authority figure in churches everywhere was in Rome and they focused on the power of money and how it had influence on others.  It caused people to question whether they were truly following their belief system. \\
\bottomrule \vspace{-2.5mm} \\
2 & Plants make oxygen which humans breathe, and they take in carbon-dioxide which humans exhale (that is, breathe out). & Plants create oxygen, of which people use and breathe in.  Then people breathe out carbon dioxide which plants need.  It is a transactional process. \\
\midrule
& Plants make oxygen which humans breathe, and they take in carbon-dioxide which humans exhale (that is, breathe out). Plants make their food from the sun by photosynthesis. They also provide shade. We make our houses from plants and make clothes from plants. Most foods that we eat are plants. Without plants, animals could not survive. & Plants create oxygen of which people breathe in.  Then when people breathe out, they are letting out carbon dioxide which plants need.  It’s a transactional process.  Plants rely on the sun where it helps them produce their own food, and this process is called photosynthesis.  Plants also provide shades, housing, materials, clothes, and even most food come from plants.  Animals cannot survive without plants. \\
\bottomrule \vspace{-2.5mm} \\
    \end{tabular}
    \label{tab:human5}
\end{table*}

\clearpage
\begin{table*}[]
    \centering
    \fontsize{8.4}{10}\selectfont
    \begin{tabular}{cp{7cm}p{7cm}}
        \toprule
        \textbf{signer} & \textbf{reference} & \textbf{human prediction} \\
\bottomrule \vspace{-2.5mm} \\
3 & They begin as funnels descending from storm clouds, and become "tornadoes" when they touch the ground. & A tornado starts when a wind cloud funnels downward until it touches the ground. Once on ground, it moves forward. \\
\midrule
& A tornado is a spinning column of very low-pressure air, which sucks the surrounding air inward and upward. They generate high winds (often 100-200 miles/hour) and can lift heavy objects into the air, carrying them as the tornado moves. They begin as funnels descending from storm clouds, and become "tornadoes" when they touch the ground. & A tornado is a funnel of low pressure air where air from outside of the funnel is sucked in and channels upward, creating the spin. A tornado forms high wind of around 100-200 mph with heavy items being pulled up into the funnel. A tornado starts when a wind cloud forms a funnel and makes its way downward onto the ground. Once it touches the ground, it moves forward and we have a tornado. \\
\bottomrule \vspace{-2.5mm} \\
3 & The setting might be an historic old building with antique furnishings, manicured grounds and a swimming pool. & The setting may be an old historical building with antique furniture. The yard may have clean cut grass and a swimming pool. \\
\midrule
& In developed countries today, providing deluxe bed and breakfasts has been raised to a sort of art-form. At the top end, B\&Bs obviously compete mainly on two main things: bedding and breakfast. Accordingly, at the finest such establishments one is apt to find the most luxurious bedding, maybe a handmade quilt or an antique bed. Breakfast may include seasonal delights of the region or the host's speciality dish. The setting might be an historic old building with antique furnishings, manicured grounds and a swimming pool. & The country has a strong economy that now provides a Bed and Breakfast that eventually became an art form. Their competitive advantage is that they provide a bed that they already have but also a breakfast. These are fancy places where you may find luxurious, antique beds with hand sewn quilts. Breakfast may be within their food season and the host may cook their specialized dish. The setting may be an old historical building with antique furniture. The yard may have clean cut grass and a swimming pool. \\
\bottomrule \vspace{-2.5mm} \\
3 & Chiao and Sharipov reported being a safe distance from the attitude adjustment thrusters. & Chiao and Sharipok were informed that during attitude adjustment that it’s best for safety reasons to step back away from the throwers. \\
\midrule
& The station maintained its attitude, despite the loss of a gyroscope earlier in the space station mission, until the end of the spacewalk. Chiao and Sharipov reported being a safe distance from the attitude adjustment thrusters. Russian ground control activated the jets and normal attitude of the station was regained. & The Space Station maintained their altitude position all throughout to the end of the space walk.  It didn't matter if they lost the gigatode in the earlier stages of the space mission.  Chiao and Sharipok were informed that during the altitude adjustments, it was best for safety reasons to back away from the thrusters.  Russian grounds control the activity jets and the station was adjusted back to their normal altitude. \\
\bottomrule \vspace{-2.5mm} \\
3 & Bird flu, or more formally avian influenza, can infect both birds and mammals. & The bird flu, formally known as the Avian influenza, affects both birds and mammals. \\
\midrule
& Bird flu, or more formally avian influenza, can infect both birds and mammals. Fewer than a thousand cases have ever been reported in humans, but some of them have been fatal. Most have involved people who work with poultry, but there is also some risk to birdwatchers. & The bird flu, formally known as the Avian Influenza, can affect both birds and mammals.  Under 1000 cases were reported in humans and some of them died.  Most of the people who contracted the disease worked with poultry but bird watchers also are at risk. \\
\bottomrule \vspace{-2.5mm} \\
3 & On the accordion, to get extra volume, you use the bellows with more pressure or speed. & The volume of the accordion is turned up by increasing the speed of which the accordion is pressed. \\
\midrule
& Make sure your hand is as relaxed as possible while still hitting all the notes correctly - also try not to make much extraneous motion with your fingers. This way, you will tire yourself out as little as possible. Remember there's no need to hit the keys with a lot of force for extra volume like on the piano. On the accordion, to get extra volume, you use the bellows with more pressure or speed. & Make sure that when you are pressing the buttons on the accordion, make sure you are relaxed and pressing while in a comfortable position.  Pressing with intensity and high speed causes premature fatigue and it is not necessary, just like while you are playing on the piano.  Also, the volume of the accordion can be turned up by increasing the speed of which you are pressing the accordion. \\
\bottomrule \vspace{-2.5mm} \\
3 & While immigration check is usually absent or a formality when you arrive in your homeland, customs control can be a hassle. & When you arrive in your homeland, immigration customs will inspect you.  Sometimes they are casual while at other times they are rigid and follow protocol.  Customs regulations can be a nuisance. \\
\midrule
& While immigration check is usually absent or a formality when you arrive in your homeland, customs control can be a hassle. Make sure you know what you can and cannot bring in and declare anything over the legal limits. & When you arrive in your homeland, immigration will do checks where it is sometimes casual but at other times more of a formality.  Customs regulations can be a nuisance.  Just make sure you know what you can and can’t bring and announce anything that is above the legal limit. \\
\bottomrule \vspace{-2.5mm} \\
    \end{tabular}
    \label{tab:human6}
\end{table*}

\clearpage
\begin{table*}[]
    \centering
    \fontsize{8.4}{10}\selectfont
    \begin{tabular}{cp{7cm}p{7cm}}
        \toprule
        \textbf{signer} & \textbf{reference} & \textbf{human prediction} \\
\bottomrule \vspace{-2.5mm} \\
3 & During the summer, also watch out for the Nordic mosquitoes. Although they do not transmit any diseases, they can be irritating. & During the summer, be aware of Nordic mosquitoes.  They don’t spread diseases but they are a pest. \\
\midrule
& Crime, and ill-will toward foreigners in general, is virtually unknown in Greenland. Even in the towns, there are no "rough areas." Cold weather is perhaps the only real danger the unprepared will face. If you visit Greenland during cold seasons (considering that the further north you go, the colder it will be), it is essential to bring warm enough clothing. The very long days in the summer can lead to problems getting sufficient sleep and associated health issues. During the summer, also watch out for the Nordic mosquitoes. Although they do not transmit any diseases, they can be irritating. & Crime towards foreigners in general never happens in Greenland, regardless of different cities not having any “rough” areas.  Cold is the only real danger for those who are unprepared to face it.  If you visit Greenland during the cold seasons, the north is dangerously frigid and it’s important to bring essential items such as warm clothing.  In the summer months, the days are long and it leads to lack of sleep which is also connected to health issues.  Be wary in the summer months of the Nordic mosquitoes.  They don’t spread diseases but they are truly a nuisance. \\
\bottomrule \vspace{-2.5mm} \\
3 & But the Royal Navy was still much stronger than the German Navy (“Kriegsmarine”) and could have destroyed any invasion fleet sent across the English Channel. & The Royal Navy is stronger than the German navy, called the Kriegsmarine.  They can defeat and overpower any invasion sent across the English channel. \\
\midrule
& With the battle for France over, Germany began to get ready to invade the island of Britain. Germany code-named the attack “Operation Sealion”. Most of the British Army's heavy weapons and supplies had been lost when it evacuated from Dunkirk, so the army was fairly weak. But the Royal Navy was still much stronger than the German Navy (“Kriegsmarine”) and could have destroyed any invasion fleet sent across the English Channel. However, very few Royal Navy ships were based near the likely invasion routes as the admirals were afraid they would be sunk by German air attack. & After the German battle against France, they were ready to overtake the British Isle. The German army’s code name for the takeover was Operation Sea Lion. Most British armies had heavy weaponry lost after their evasion from Dunkirk so they were weakened.  The royal navy is still stronger than the German navy, called the Kriegsmarine.  They can destroy and defeat any invasions sent across the English Channel.  However, where there were a few royal navy ships near the location for planned takeover, officers feared Germans would fly over and sink those ships. \\
\bottomrule \vspace{-2.5mm} \\
4 & However, this shouldn't really be off your concern, because often tourists are shuffled around to fill the cars. & Really most trips tend to be a collection of cars, vans, buses and others. \\
\midrule
& Tours are cheaper for larger groups, so if you're by yourself or with just one friend, try to meet other people and form a group of four to six for a better per-person rate. However, this shouldn't really be off your concern, because often tourists are shuffled around to fill the cars. It seems actually to be more a way of tricking people into believing they have to pay more. & Today I am going to talk about tours. Oftentimes, if you go on a tour as a group, they will give you a discounted rate rather than if you were alone or as a couple, it costs more. If there were four or six of us, we can all chip in on the cost and it’s cheaper. In the process, you may get a chance to meet other people too. Most tours tend to include cars, vans, buses and others of that source. The concept influenced me to think that being alone is more expensive and therefore tours alone are not necessary. The tour industry has influenced me to believe that. \\
\bottomrule \vspace{-2.5mm} \\
4 & If you want to see the world on the cheap, for necessity, lifestyle or challenge, there are some ways to do that. & Suppose you wanted to travel and see the world on a low budget, whether you chose lifestyle or facing challenges along the way, there are different ways to do it. \\
\midrule
& If you want to see the world on the cheap, for necessity, lifestyle or challenge, there are some ways to do that. Basically, they fall into two categories: Either work while you travel or try and limit your expenses. This article is focused on the latter. For those willing to sacrifice comfort, time and predictability to push expenses down close to zero, see minimum budget travel. The advice assumes that travellers do not steal, trespass, participate in the illegal market, beg, or otherwise exploit other people for their own gain. & Suppose you wanted to travel and see the world on a low budget, whether you chose between lifestyle or face challenges, there are different ways to do it. There are basically two ways to do it. There is work-study travel and that will limit your spending. The second way to do it, I will go into depth later. So, for those of you who are willing to sacrifice the luxury of comfort and a predictable schedule, you can reduce your expenditure to be very low and can travel on a minimal travel budget. My advice to you is don’t steal, don’t trespass on property, don’t participate in the illegal market, don’t beg or don’t take advantage of others. For your own good, just don’t. \\
\bottomrule \vspace{-2.5mm} \\
    \end{tabular}
    \label{tab:human7}
\end{table*}

\clearpage
\begin{table*}[]
    \centering
    \fontsize{8.4}{10}\selectfont
    \begin{tabular}{cp{7cm}p{7cm}}
        \toprule
        \textbf{signer} & \textbf{reference} & \textbf{human prediction} \\
\bottomrule \vspace{-2.5mm} \\
4 & Protests also took place in Paris, Sofia in Bulgaria, Vilnius in Lithuania, Valetta in Malta, Tallinn in Estonia, and Edinburgh and Glasgow in Scotland. & Protests also happened at other locations in Paris, Sofia in Bulgaria, Vilnius in Lithuania, Valletta in Malta, Tallinn in Estonia, and Edinburgh as well as Glasgow in Scotland.  There were many locations. \\
\midrule
& Organisers of the protest said about 100,000 people turned up in German cities such as Berlin, Cologne, Hamburg, and Hanover. In Berlin, police estimated 6,500 protestors. Protests also took place in Paris, Sofia in Bulgaria, Vilnius in Lithuania, Valetta in Malta, Tallinn in Estonia, and Edinburgh and Glasgow in Scotland. In London, about 200 people protested outside some major copyright holders' offices. Last month, there were major protests in Poland when that country signed ACTA, which has led to the Polish government deciding not to ratify the agreement, for now. Latvia and Slovakia have both delayed the process of joining ACTA. & Organizations of the protests said about 100,000 people showed up in four different German towns, which are Berlin, Cologne, Hamburg, and Hanover.  In Berlin, the police estimated about 2,500 protesters.  There were protests happening in other locations too.  Other locations include Paris, Sofia in Bulgaria, Vilnius in Lithuania, Valletta in Malta, Tallinn in Estonia, and Edinburgh as well as Glasgow in Scotland.  There were many locations.  In London, around 200 people protested there.  Outside of some big copyright holder offices where there were several larger protests last month.  In Poland, when that country signed ACTA, that caused other governments to be indecisive about signing an agreement. Latvia and Slovakia also had a delayed response in the process of joining ACTA. \\
\bottomrule \vspace{-2.5mm} \\
4 & "We have a year-long financial crisis, which has had its most acute moment in the past two months, and I think now the financial markets are beginning to recover." & In the past year, there was a long financial crisis.  The past two months were the worst but I think we are finally beginning to recover. \\
\midrule
& "Regarding the global financial situation, Zapatero continued by saying that "the financial system is a part of the economy, a crucial part. We have a year-long financial crisis, which has had its most acute moment in the past two months, and I think now the financial markets are beginning to recover." & On the global financial situation, Zapatero said the financial system is a very important part of the economic system.  The past year was a long financial crisis, especially the past two months.  I think we are finally beginning to recover. \\
\bottomrule \vspace{-2.5mm} \\
4 & Ten years later, he led the Soviet part of the Apollo–Soyuz mission symbolizing that the Space Race was over. & Ten years later, they were leading as shown in the Soviet Apollo Soyuz mission, which was representative of how their space race process worked effectively. \\
\midrule
& Leonov, also known as "cosmonaut No. 11", was part of the Soviet Union's original team of cosmonauts. On March 18, 1965, he performed the first manned extravehicular activity (EVA), or "spacewalk", remaining alone outside the spacecraft for just over twelve minutes. He received the "Hero of the Soviet Union", the Soviet Union's highest honor, for his work. Ten years later, he led the Soviet part of the Apollo–Soyuz mission symbolizing that the Space Race was over. & Leonov is a well known as cosmonaut \#11 who is part of the Soviet Union’s original team for Astronauts.  On March 18, 1965, he was the first person to be involved with an extravehicular activity, EVA, where he went into outer space and did a spacewalk for 12 minutes before he returned.  He was recognized as a hero in the Soviet Union and received highest honors for his work.  Ten years later, he led the Soviet Apollo Soyuz mission, demonstrative of their space race process and how it worked smoothly and effectively. \\
\bottomrule \vspace{-2.5mm} \\
4 & However, in June 1956, Krushchev's promises were put to the test when riots in Poland, where workers were protesting against food shortages and wage cuts, turned into a general protest against Communism. & In the month of June in 1956, Krush Chev made promises which caused disruptions because people and workers in Poland were revolting due to food shortage and low earnings. It resulted in a revolt against communism. \\
\midrule
& However, in June 1956, Krushchev's promises were put to the test when riots in Poland, where workers were protesting against food shortages and wage cuts, turned into a general protest against Communism. Although in the end, Krushchev sent in tanks to restore order, he did give way to some economic demands and agreed to appoint the popular Wladyslaw Gomulka as the new prime minister. & In June of 1956, Khrushchev made empty promises to his people that resulted in disruptions.  People in Poland were protesting due to food shortages and low earnings and that also subsequently resulted in a revolt against communism.  So in the midst of the revolts, Khrushchev deployed military and battle tanks to restore order and control in the people.  He consented to demands from the people for economic stability, food and increased earnings, and also appointed a popular prime minister named Wladyslaw Gomulka. \\
\bottomrule \vspace{-2.5mm} \\
4 & However, Charles went to university at Trinity College, Cambridge where he studied Anthropology and Archaeology, and later History, earning a 2:2 (a lower second class degree). & However, Charles went to college at Trinity College in Cambridge.  He studied Anthropology and Archaeology.  He also eventually studies History.  He earned a 2.2 score, which is really low, so he earned a second class degree. \\
\midrule
& Traditionally, the heir to the throne would go straight into the military after finishing school. However, Charles went to university at Trinity College, Cambridge where he studied Anthropology and Archaeology, and later History, earning a 2:2 (a lower second class degree). Charles was the first member of the British Royal Family to be awarded a degree. & Tradition of the heir to the throne is you go to school and then you immediately go to the military. However, Charles went to university at Trinity College in Cambridge where he studied Anthropology and Archaeology there.  He also eventually studied History as well.  There he earned a 2.2 score, which is very low, and is a second class degree.  Charles is the first member of the British royal family to have been awarded a degree. \\
\bottomrule \vspace{-2.5mm} \\
    \end{tabular}
    \label{tab:human8}
\end{table*}

\clearpage
\begin{table*}[t]
    \centering
    \fontsize{8.4}{10}\selectfont
    \begin{tabular}{cp{7cm}p{7cm}}
        \toprule
        \textbf{signer} & \textbf{reference} & \textbf{human prediction} \\
\bottomrule \vspace{-2.5mm} \\
4 & He recently lost against Raonic in the Brisbane Open. & Recently, there was a loss against Raonic in the Brisbane Opening. \\
\midrule
& Nadal's head to head record against the Canadian is 7–2. He recently lost against Raonic in the Brisbane Open. Nadal bagged 88\% net points in the match winning 76 points in the first serve. After the match, King of Clay said, "I am just excited about being back in the final rounds of the most important events. I am here to try to win this. & Nadal’s record against Canada in tennis was 7-2. The recent loss against Raonic in the Brisbane opening was a disappointment. Nadal impacted 88\% of the points in the match since being awarded 76 points. After the game was over, the King of Clay said he was excited to return to watch the last round of the most important game event. He said a win is a must! \\
\bottomrule \vspace{-2.5mm} \\
    \end{tabular}
    \label{tab:human9}
\end{table*}



\end{document}